\documentclass{article}

\usepackage[authoryear, comma]{natbib}

     \usepackage[preprint]{neurips_2021}

\usepackage[utf8]{inputenc} 
\usepackage[T1]{fontenc}    
\usepackage{hyperref}       
\usepackage{url}            
\usepackage{booktabs}       
\usepackage{amsfonts}       
\usepackage{nicefrac}       
\usepackage{microtype}      
\usepackage{xcolor}         

\usepackage{amsmath}        
\usepackage{caption}
\usepackage{subcaption}
\usepackage{graphicx}

\newcommand\boldx{\mathbf{x}}
\newcommand\boldz{\mathbf{z}}

\newcommand\zpres{\boldz^{\mathsf{pres}}}
\newcommand\zcat{\boldz^{\mathsf{cat}}}
\newcommand\zwhat{\boldz^{\mathsf{what}}}
\newcommand\zwhere{\boldz^{\mathsf{where}}}
\newcommand\zdepth{\boldz^{\mathsf{depth}}}
\newcommand\zlocal{\boldz^{\mathsf{local}}}
\newcommand\zavg{\boldz^{\mathsf{avg}}}

\newcommand\ep{\mathbb{E}}

\title{GMAIR : Unsupervised Object Detection Based on Spatial Attention and Gaussian Mixture}

\author{
  Weijin Zhu, Yao Shen, Linfeng Yu, Lizeth Patricia Aguirre Sanchez \\
  Department of Computer Science \\
  Shanghai Jiao Tong University \\
  \texttt{\{weijinzhu,yshen,ylf2017,lizethaguirre\}@sjtu.edu.cn} \\
}

\begin{document}

\nocite{*}

\maketitle



\begin{abstract}
  Recent studies on unsupervised object detection based on spatial attention have achieved promising results. Models, such as AIR and SPAIR, output ``what'' and ``where'' latent variables that represent the attributes and locations of objects in a scene, respectively. Most of the previous studies concentrate on the ``where'' localization performance; however, we claim that acquiring ``what'' object attributes is also essential for representation learning. This paper presents a framework, GMAIR, for unsupervised object detection. It incorporates spatial attention and a Gaussian mixture in a unified deep generative model. GMAIR can locate objects in a scene and simultaneously cluster them without supervision. Furthermore, we analyze the ``what'' latent variables and clustering process. Finally, we evaluate our model on MultiMNIST and Fruit2D datasets and show that GMAIR achieves competitive results on localization and clustering compared to state-of-the-art methods.
\end{abstract}

\section{Introduction}

The perception of human vision is naturally hierarchical. We can recognize objects in a scene at a glance and classify them according their appearances, functions, and other attributes. It is expected that an intelligent agent can also decompose scenes to meaningful object abstraction, which is known as an object detection task in machine learning. In the last decade, there have been significant developments in supervised object detection tasks. However, its unsupervised counterpart continues to be challenging.

Recently, there has been some progress in unsupervised object detection. Attend, infer, repeat (AIR, \cite{eslami2016attend}), which is a variational autoencoder (VAE, \cite{kingma2013auto}) based method, achieved encouraging results. Spatially invariant AIR (SPAIR, \cite{crawford2019spatially}) replaced the recurrent network in AIR by a convolutional network that attained better scalability and lower computational cost. SPACE (\cite{lin2020space}), which combines spatial-attention and scene-mixture approaches, performed better in background prediction.

Despite the recent progress in unsupervised object detection, results of previous studies remain unsatisfactory. One of the reasons for this could be that previous studies on unsupervised object detection were mainly concentrated on object localization and lacked analysis and evaluation on the ``what'' latent variables, which represent the attributes of objects. These variables are essential for many tasks such as clustering, image generation, and style transfer.
Another important concern is that they do not directly reason about the category of objects in the scene, which is beneficial to know in many cases, unlike most of the studies on corresponding supervised tasks.

This paper presents a framework for unsupervised object detection that can directly reason about the category and localization of objects in the scenes and provide an intuitive way to analyze the ``what'' latent variables by simply incorporating a Gaussian mixture prior assumption.
In Sec. \ref{sec:gmair}, we introduce the architecture of our framework, GMAIR.
We introduce related works in Sec. \ref{sec:relatedworks}.
We analyze the ``what'' latent variables in Sec. \ref{sec:what}. We describe our model for image generation in Sec. \ref{sec:imagegeneration}. Finally, we present quantitative evaluation results of both clustering and localization in Sec. \ref{sec:eval}.

Our main contributions are:
\begin{itemize}
  \item We combine spatial attention and a Gaussian mixture in a unified deep generative model, enabling our model to cluster discovered objects.
  \item We analyze the ``what'' latent variables, which are essential because they represent the attributes of the objects.
  \item Our method achieves competitive results on both clustering and localization compared to state-of-the-art methods.
\end{itemize}

\section{Gaussian Mixture Attend, Infer, Repeat}
\label{sec:gmair}

\begin{figure}
  \centering

  \def\svgwidth{\columnwidth}
  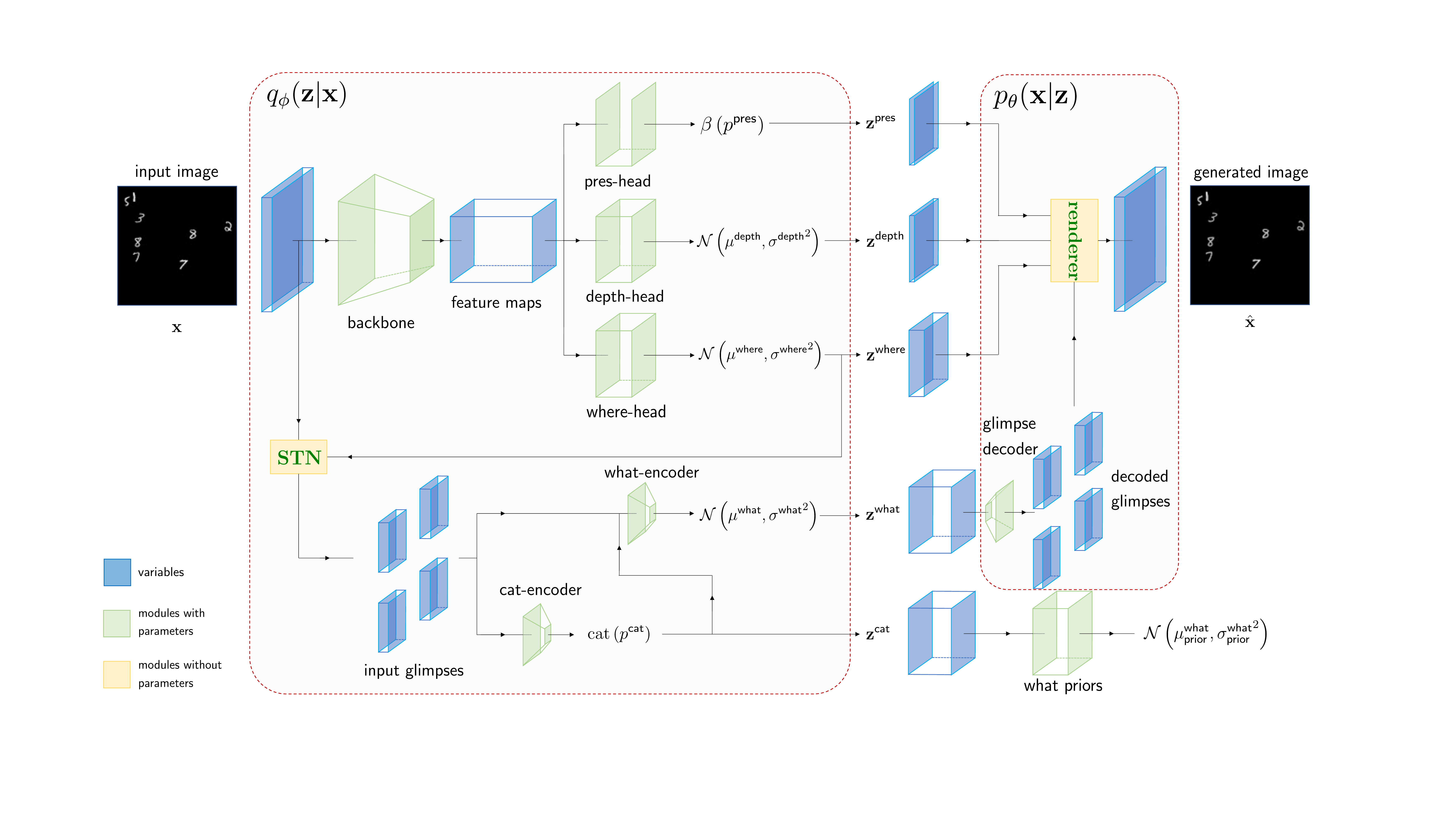

  \caption{
  Architecture of GMAIR.
  This is a VAE-based model that consists of a probabilistic encoder, $q_\phi(\boldz|\boldx)$, and a probabilistic decoder, $p_\theta(\boldx|\boldz)$.
  In encoder $q_\phi(\boldz|\boldx)$, feature maps with dimension $H\times W\times D$ are extracted from data $\boldx$ going through a backbone network representing feature of $H\times W$ divided regions. They are then fetched into three separated modules: pres-head, depth-head, and where-head, which produce the posterior of $\zpres$, $\zdepth$, and $\zwhere$, respectively. A cat-encoder module generates The posterior of $\zcat$ with $H\times W$ input glimpses transformed by a spatial transformer network (STN) as input, and the posterior of $\zwhat$ is generated by a what-encoder module with $H\times W$ input glimpses and $\zcat$ as input.
  In decoder $p_\theta(\boldx|\boldz)$, each $H\times W$ latent $\zwhat$ is fetched into a glimpse decoder to generate decoded glimpses rendered by the renderer to recover to the final generated image.
  Finally, the priors of $\zpres, \zdepth, \zwhere$, and $\zcat$ are fixed, whereas the prior of $\zwhat$ is generated by a ``what priors'' module using $\zcat$ as input.
  }
  \label{fig:gmair}
\end{figure}

In this section, we introduce our framework, GMAIR, for unsupervised object detection. GMAIR is a spatial-attention model with a Gaussian mixture prior assumption for the ``what'' latent variables, and this enables the model to cluster discovered objects. An overview of GMAIR is presented in Fig. \ref{fig:gmair}.

\subsection{Structured Object-semantic Latent Representation}
\label{sec:latent}

We follow SPAIR to attain object abstraction latent variables (\cite{crawford2019spatially}); the image is divided into $H\times W$ regions. Latent variables $\boldz = (\boldz_{1}, \boldz_{2}, ..., \boldz_{HW})$ is a concatenation of $HW$ latent variables where $\boldz_{i}$ is the latent variable for the $i$-th region representing the semantic feature of the object centered in the $i$-th region.
Furthermore, for each region we divide $\boldz_{i}$ into five seperate latent variables, $\boldz_{i} = (\zpres_{i}, \zwhat_{i}, \zcat_{i}, \zwhere_{i}, \zdepth_{i})$,
where $\zpres_{i} \in \{0,1\}, \zwhat_{i} \in \mathbb{R}^{A}, \zcat_{i} \in \{0,1\}^{C}, \zwhere_{i} \in \mathbb{R}^4, \zdepth_{i} \in \mathbb{R}$, $A$ is the dimension of ``what'' latent variables and $C$ is the number of clusters.
The meaning of $\zpres,\zwhat,\zwhere,\zdepth$ are the same as in \cite{crawford2019spatially}, while $\zcat$ are one-hot vectors for categories.

GMAIR imposes a prior on those latent variables as follow:

\begin{equation} \label{eqn:pz}
  p(\boldz) = \prod_{i=1}^{HW} p(\zpres_i)\left(p(\zcat_i)p(\zwhat_i|\zcat_i)p(\zwhere_i)p(\zdepth_i)\right)^{\zpres_i}.
\end{equation}

\paragraph{Gaussian Mixture Prior Assumption}
Latent variables $\zcat$ are one-hot vectors that act as classification indicators. They obey the categorical distribution, $\operatorname{Cat}(\boldsymbol{\pi})$, where $\boldsymbol{\pi} \in [0,1]^{C}$. For simplicity, we assume that $\boldsymbol{\pi}_k=1/C$ for all $1\leq k \leq C$.

We assume that $\zwhat_i$ conditional on $\zcat_i$ obeys a Gaussian distribution. In that case, $\zwhat_i$ obeys a Gaussian mixture model, that is,

\begin{align}
  p(\zwhat_i) & = \sum_{k=1}^{C} p(\zcat_{i,k}=1) p(\zwhat_i | \zcat_{i,k}=1) \\
    & = \sum_{k=1}^{C} p(\zcat_{i,k}=1) f(x ; \mu_k, \sigma^2_k) \nonumber
\end{align}

where $f(x;\mu,\sigma^2) = \frac{1}{\sigma\sqrt{2\pi}}\exp\left(\frac{(x-\mu)^2}{2\sigma^2}\right)$ is the probability density function of Gaussian distribution, $\mu_k, \sigma_k$ ($k=1..C$) are the mean and standard derivation of the $k$-th Gaussian distribution. We let $\mu_k$ and $\sigma_k$ be learnable parameters that are jointly trained with other parameters.
During the implementation, $\mu_k = \mu(\zcat_i)$ and $\sigma_k = \sigma(\zcat_i)$ if $\zcat_{i,k} = 1$ where $\mu$ and $\sigma$ can be modeled as linear layers. They are called ``what priors'' module in Figure \ref{fig:gmair}.

For other latent variables, $\zpres$ are modeled using a Bernoulli distribution, $\beta(p)$, where $p$ is the present probability.
$\zwhere$ and $\zdepth$ are modeled using normal distributions,
$\mathcal{N}(\mu^{\mathsf{where}}_{\mathsf{prior}}, {\sigma^{\mathsf{where}}_{\mathsf{prior}}}^2)$
and $\mathcal{N}(\mu^{\mathsf{depth}}_{\mathsf{prior}}, {\sigma^{\mathsf{depth}}_{\mathsf{prior}}}^2)$, respectively.
All priors of latent variables are listed in Table \ref{table:priors}.

\begin{table}
  \caption{Priors of latent variables}
  \label{table:priors}
  \centering
  \begin{tabular}{ll}
    \toprule
    Latent Variables     & Priors     \\
    \midrule
    $\zpres$ & $\beta(p)$ \\
    $\zwhat$ &  $\mathcal{N}(\mu(\zcat), {\sigma(\zcat_i)}^2)$ \\
    $\zcat$ & $\operatorname{Cat}(\boldsymbol{\pi})$ \\
    $\zwhere$ & $\mathcal{N}(\mu^{\mathsf{where}}_{\mathsf{prior}}, {\sigma^{\mathsf{where}}_{\mathsf{prior}}}^2)$ \\
    $\zdepth$ & $\mathcal{N}(\mu^{\mathsf{depth}}_{\mathsf{prior}}, {\sigma^{\mathsf{depth}}_{\mathsf{prior}}}^2)$ \\
    \bottomrule
  \end{tabular}
\end{table}

\subsection{Inference and Generation Model}

\paragraph{Inference Model $q_\phi(\boldz|\boldx)$}
In the inference model, latent variables conditional on data $\boldx$ are modeled by Eqn. \ref{eqn:qzx}.

\begin{equation}\label{eqn:qzx}
  q(\boldz | \boldx) = \prod_{i=1}^{HW} q(\zpres_{i}|\boldx) \left( q(\zwhere_{i}|\boldx) q(\zdepth_{i}|\boldx) q(\zcat_{i}|\boldx, \zwhere_{i}) q(\zwhat_{i}|\boldx, \zwhere_{i}, \zcat_{i}) \right)^{\zpres_i}.
\end{equation}

During implementation, feature maps with dimension $H\times W\times D$ are extracted from a backbone network using data $x$ as input, where $D$ is the number of channels of feature maps. Further, the posteriors of $\zpres, \zwhere$, and $\zdepth$ are reasoned by pres-head, where-head, and depth-head, respectively. Input images are cropped into $H\times W$ glimpses by a spatial transformer network, and each of these is transferred to the cat-encoder module to generate posteriors of $\zcat$. Subsequently, we use the concatenation of the $i$-th glimpse and $\zcat_i$ $(1\leq i\leq HW)$ as the input of the what-encoder to generate posteriors of $\zwhat$.

\paragraph{Generation Model $p_\theta(\boldx|\boldz)$}

In the generation model, each $\zwhat_i$ $(1\leq i\leq HW)$ is changed back into a glimpse by using a glimpse decoder. Then, a renderer combines $HW$ glimpses to generate $\hat{\boldx}$. We use the same render algorithm as in previous studies (\cite{eslami2016attend, crawford2019spatially}).

\subsection{The Loss Functions}
\paragraph{Evidence Lower Bound}
In general, we learn parameters of VAE jointly by maximizing the evidence lower bound (ELBO), which can be formulated as:

\begin{align}\label{eqn:elbo}
  ELBO & = \ep_{q(\boldz|\boldx)} \left[ \log \left( \frac{p(\boldx,\boldz)}{q(\boldz|\boldx)} \right) \right] \\
    {} & = \ep_{q(\boldz|\boldx)} \left[ \log \left( p(\boldx|\boldz) \right) \right] - \ep_{q(\boldz|\boldx)} \left[ \log \left( \frac{q(\boldz|\boldx)}{p(\boldz)} \right) \right] \nonumber
\end{align}

where, the first term is called the reconstruction term denoted by $-L_{\mathsf{recon}}$ and the second term, the regularization term. The regularization term can be further decomposed into five terms by substituting Eqn. \ref{eqn:pz} and Eqn. \ref{eqn:qzx} into Eqn. \ref{eqn:elbo}, and each of the five terms corresponding to the Kullback--Leibler divergence (or its expectation) between a type of latent variables and its prior:
\begin{equation}\label{eqn:kl}
  \ep_{q(\boldz|\boldx)} \left[ \log \left( \frac{q(\boldz|\boldx)}{p(\boldz)} \right) \right] =
  L_{\mathsf{pres}} + L_{\mathsf{where}} + L_{\mathsf{depth}} + L_{\mathsf{cat}} + L_{\mathsf{what}}.
\end{equation}
The terms in Eqn. \ref{eqn:kl} are:
\begin{align}\label{eqn:allkl}
  L_{\mathsf{pres}}\ \ = & \sum_{i=1}^{HW} KL \left(q(\zpres_i|\boldx) || p(\zpres_i) \right)  \\
  L_{\mathsf{where}} = & \sum_{i=1}^{HW} q(\zpres_i=1|\boldx) KL \left( q(\zwhere_i|\boldx) || p(\zwhere_i) \right)  \\
  L_{\mathsf{depth}} = & \sum_{i=1}^{HW} q(\zpres_i=1|\boldx) KL \left( q(\zdepth_i|\boldx) || p(\zdepth_i) \right)  \\
  L_{\mathsf{cat}}\ \ \ = & \sum_{i=1}^{HW} q(\zpres_i=1|\boldx) \ep_{q(\zwhere_i|\boldx)} \left[ KL \left( q(\zcat_i|\boldx,\zwhere_i) || p(\zcat_i) \right) \right] \\
  L_{\mathsf{what}}\ = & \sum_{i=1}^{HW} q(\zpres_i=1|\boldx) \ep_{q(\zwhere_i, \zcat_i|\boldx)} \left[ KL \left( q(\zwhat_i|\boldx,\zwhere_i,\zcat_i) || p(\zwhat_i|\zcat_i) \right) \right].
\end{align}

A complete derivation is given in Appendix \ref{appendix:elbo}.

\paragraph{Overlap Loss}

During actual implementation, we find that penalizing on overlaps of objects sometimes helps. Therefore, we introduce an auxiliary loss called overlap loss.
First, we calculate $HW$ images with size $3\times H^{\mathsf{img}} \times W^{\mathsf{img}}$, where $H^{\mathsf{img}}$ and $W^{\mathsf{img}}$ are respectively the height and width of the input image, transformed by $HW$ decoded glimpses by a spatial transformer network.
The overlap loss is then calculated as the average of the sum subtract by the maximum for each $H^{\mathsf{img}}\times W^{\mathsf{img}}$ pixels.

This loss, inspired by the boundary loss in SPACE (\cite{lin2020space}), is utilized to penalize if the model tries to split a large object into multiple smaller ones. However, we achieve this by using a different calculation method that incurs a lower computational cost.

\paragraph{Total Loss}
The total loss is:
\begin{equation}
  L = \sum_{x\in S} \alpha_{x} L_{x}
\end{equation}
where, $S=\left\{ \mathsf{recon}, \mathsf{overlap}, \mathsf{pres}, \mathsf{where}, \mathsf{depth}, \mathsf{cat}, \mathsf{what} \right\}$, and $\alpha_{:}$ are the coefficients of the corresponding loss terms.

\section{Related Works}
\label{sec:relatedworks}

Several studies on unsupervised object detection have been conducted, including spatial-attention methods such as AIR (\cite{eslami2016attend}), SPAIR (\cite{crawford2019spatially}), and SPACE (\cite{lin2020space}),
and scene-mixture methods such as MONet (\cite{burgess2019monet}), IODINE (\cite{greff2019multi}), and GENESIS (\cite{engelcke2019genesis}). Most of them including our work are based on a VAE (\cite{kingma2013auto}).

The AIR (\cite{eslami2016attend}) framework uses a VAE-based hierarchical probabilistic model marking a milestone in unsupervised scene understanding. In AIR, latent variables are structured into groups of latent variables $\boldz_{1:N}$, for $N$ discovered objects, each of which consists of ``what,'' ``where,'' and ``presence'' variables. A recurrent neural network is used in the inference model to produce $\boldz_{1:N}$, and there is a decoder network for decoding the ``what'' variables of each object in the generation model. A spatial transformer network (\cite{jaderberg2015spatial}) is used for rendering.

Because AIR attends one object at a time, it does not scale well to scenes that contain many objects. SPAIR (\cite{crawford2019spatially}) attempted to address this issue by replacing the recurrent network with a convolutional network that follows a spatially invariant assumption. Similar to YOLO (\cite{redmon2016you}), in SPAIR, the locations of objects are specified relative to local grid cells.

Scene-mixture models such as MONet (\cite{burgess2019monet}), IODINE (\cite{greff2019multi}), and GENESIS (\cite{engelcke2019genesis}) perform segmentation instead of explicitly finding the $\zwhere$ location of objects. SPACE (\cite{lin2020space}) employs a combination of both methods. It consists of a spatial-attention model for the foreground and a scene-mixture model for the background.

In the area of deep unsupervised clustering, recent methods include
AAE (\cite{makhzani2015adversarial}), GMVAE (\cite{dilokthanakul2016deep}), IIC (\cite{ji2019invariant}).
AAE combines the ideas of generative adversarial networks and variational inference.
GMVAE uses a Gaussian mixture model as a prior distribution.
In IIC, objects are clustered by maximizing mutual information of pairs of images.
All of them show promising results on unsupervised clustering.

GMAIR incorporates a Gaussian mixture model for clustering, similar to the GMVAE framework\footnote{We also refer to a blog post
(http://ruishu.io/2016/12/25/gmvae/) published by Rui Shu.}.
It worth noting that our attempt may simply be a choice amongst many given options.
Unless previous research, our main contribution is to show the feasibility of performing clustering and localization simultaneously. Moreover, our method provides a simple and intuitive way to analyze the mechanics of the detection process.

\section{Models and Experiments}
\label{sec:experiments}

The experiments were divided into three parts: \textbf{a)} the analysis of ``what'' representation and clustering along with the iterations, \textbf{b)} image generation, and \textbf{c)} quantitative evaluation of the models.

We evaluate the models on two datasets :
\begin{itemize}
  \item \textbf{MultiMNIST}: A dataset generated by placing 1--10 small images randomly chosen from \textbf{MNIST} (a standard handwritten digits dataset, (\cite{lecun1998mnist})) to random positions on $128\times 128$ empty images.

  \item \textbf{Fruit2D}: A dataset collected from a real-world game. In the scenes, there are $9$ types of fruits of various sizes. There is a large difference between both the number and the size of small objects and large objects. The ratio of the size of the largest type of objects to that of the smallest type of objects is \textasciitilde$6$, and there are \textasciitilde$31$ times objects in the smallest size than in the largest size. These settings make it difficult to perform localization and clustering.
\end{itemize}

In the experiments, we compared GMAIR to two models, SPAIR and SPACE, both of which achieve state-of-the-art in unsupervised object detection in localization performance. Separated Gaussian mixture models are applied to the ``what'' latent variables generated by the compared models to obtain the clustering results. We set the number of clusters $C=10$ and Monte Carlo samples $M=1$ except as otherwise defined for all experiments. We present the details of models in Appendix \ref{appendix:imp}.

It is worth mentioning that the model sometimes successfully locates an object and encloses it with a large box. In that case, IoU between the ground truth and the predicted one will be small, and therefore, will not count to be a correct bounding box when calculating AP. We fix this issue by removing the empty area in generated glimpses to obtain the real size of predicted boxes.

\subsection{``What'' Representation and Cluster Analysis}
\label{sec:what}

\begin{figure}
  \begin{subfigure}{.24\textwidth}
    \centering
    \includegraphics[width=\linewidth]{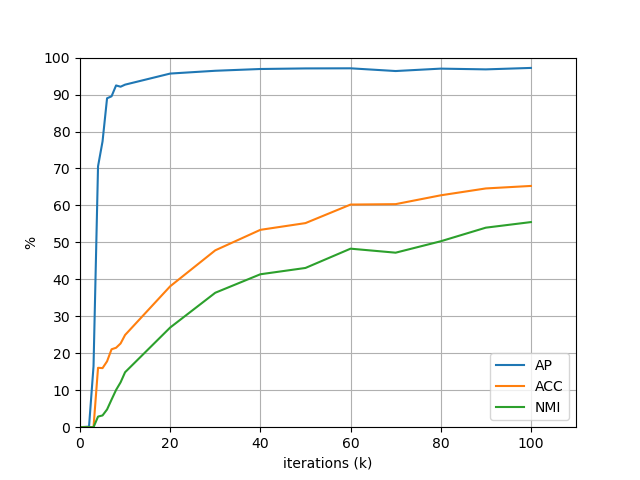}
    \caption{AP(IoU=0.5), ACC and NMI during training}
    \label{fig:what-ap}
  \end{subfigure}
  \begin{subfigure}{.24\textwidth}
    \centering
    \includegraphics[width=\linewidth]{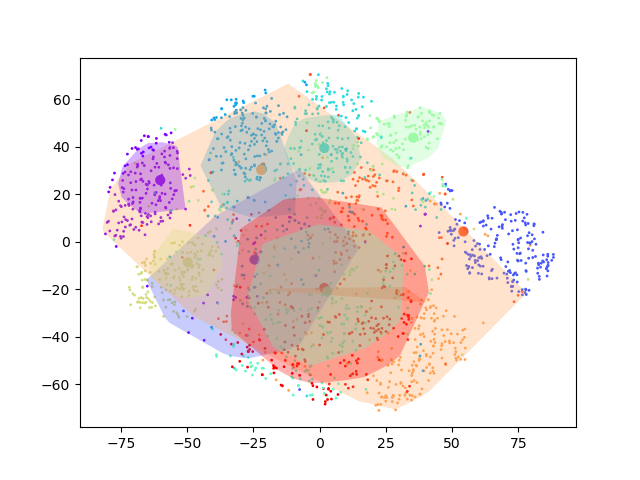}
    \caption{``What'' latent space, at 10k iterations}
    \label{fig:what-10}
  \end{subfigure}
  \begin{subfigure}{.24\textwidth}
    \centering
    \includegraphics[width=\linewidth]{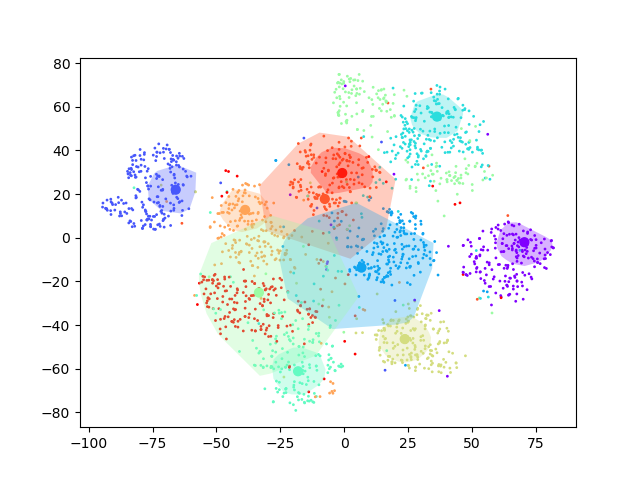}
    \caption{``What'' latent space, at 50k iterations}
    \label{fig:what-50}
  \end{subfigure}
  \begin{subfigure}{.24\textwidth}
    \centering
    \includegraphics[width=\linewidth]{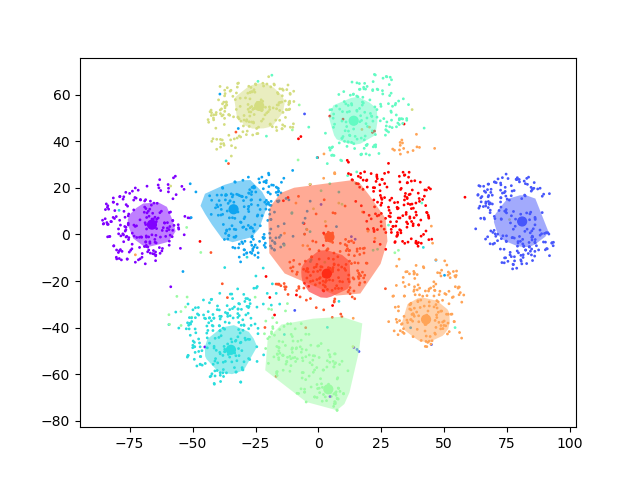}
    \caption{``What'' latent space, at 100k iterations}
    \label{fig:what-100}
  \end{subfigure}
  \newline
  \begin{subfigure}{.24\textwidth}
    \centering
    \includegraphics[width=.75\linewidth]{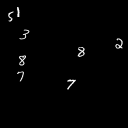}
    \caption{Original image}
    \label{fig:what-image-ori}
  \end{subfigure}
  \begin{subfigure}{.24\textwidth}
    \centering
    \includegraphics[width=.75\linewidth]{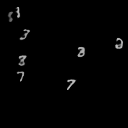}
    \caption{Generated image, at 10k iterations}
    \label{fig:what-image-10}
  \end{subfigure}
  \begin{subfigure}{.24\textwidth}
    \centering
    \includegraphics[width=.75\linewidth]{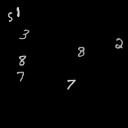}
    \caption{Generated image, at 50k iterations}
    \label{fig:what-image-50}
  \end{subfigure}
  \begin{subfigure}{.24\textwidth}
    \centering
    \includegraphics[width=.75\linewidth]{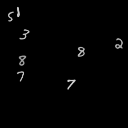}
    \caption{Generated image, at 100k iterations}
    \label{fig:what-image-100}
  \end{subfigure}
  \caption{``What'' representation and cluster analysis.
  (a) Average precision (AP), accuracy (ACC), normalized mutual information (NMI) during training.
  (b-d) Visualized ``what'' latent space by t-SNE (\cite{van2008visualizing}) at 10k, 50k, and 100k iterations, respectively. Each small dot represents a sample of $\zwhat$, and different colors represent the ground-truth categories of the corresponding objects. The large dots are $\mu_k(1\leq k\leq C)$ described in Sec. \ref{sec:latent}, and each of these can be seen as the center of a cluster.
  The closures represent results of clustering, which are closures of the closest $n$ $\zwhat$ points to $\mu_k$ that are assigned to the $k$-th cluster (where $1\leq k\leq C$ and we choose $n=200$). The color of $\mu_k(1\leq k\leq C)$ and closures are decided by a matching algorithm such that a maximum number of $\zwhat$ are correctly classified to the ground-truth label.
  (e) Sample of original image.
  (f-h) Samples of generated image at 10k, 50k, and 100k iterations, respectively.}
  \label{fig:what}
\end{figure}

We conducted the experiments using the MultiMNIST dataset. We ran GMAIR for 440k iterations and observed the change in the values of the average precision (AP) of bounding boxes, accuracy (ACC), and normalized mutual information (NMI) of clustering until 100k iterations. We also visualized the ``what'' latent variables in the latent space during the process, as shown in Fig. \ref{fig:what}. Although all values continued to increase even after 100k iterations, the visualization results were similar to those at the 100k iteration. For integrity, we reserved the results from 100k to 440k iterations in Appendix \ref{appendix:exps}. Details of calculating the AP, ACC, and NMI are discussed in Appendix \ref{appendix:ap,acc,nmi}.

The results showed that at an early stage (\textasciitilde10k iterations) of training, models can already locate objects well with AP $>0.9$ (Fig. \ref{fig:what-ap}). At the same time, $\zwhat$, representations of objects were still evolving, and the results of clustering (in Fig. \ref{fig:what-10}) was not desirable ((ACC, NMI) was ($0.24, 0.15$)); the digits were a blur in Fig. \ref{fig:what-image-10}.
After 50k and 100k iterations of training, the clustering effect of $\zwhat$ was increasingly apparent, and the digits were clearer (Fig. \ref{fig:what-image-50}, \ref{fig:what-image-100}). The clustering results ((ACC, NMI) was ($0.55, 0.43$) at 50k, and ($0.65, 0.55$) at 100k iterations) were improved (Fig. \ref{fig:what-50}, \ref{fig:what-100}).

It should be noted that even if the clustering effect of $\zwhat$ is sufficiently enough, the model may fail to locate the centers of clusters (for example, the large cluster in light red in Fig. \ref{fig:what-100}), leading to poor clustering results.
In the worst case, the model may learn to converge all $\mu_k, \sigma_k(1\leq k\leq C)$ to the same values, $\mu^{*}, \sigma^{*}$, and the Gaussian mixture model may degenerate to a single Gaussian distribution, $\mathcal{N}(\mu^{*}, {\sigma^{*}}^2)$, resulting in a miserable clustering result. In general, we found that this phenomenon usually occurs at the early stage of training and can be avoided by adjusting the learning rate of relative modules and the coefficients of the loss functions.

\subsection{Image Generation}
\label{sec:imagegeneration}
\begin{figure}
  \begin{subfigure}{.45\textwidth}
    \centering
    \includegraphics[width=0.9\linewidth]{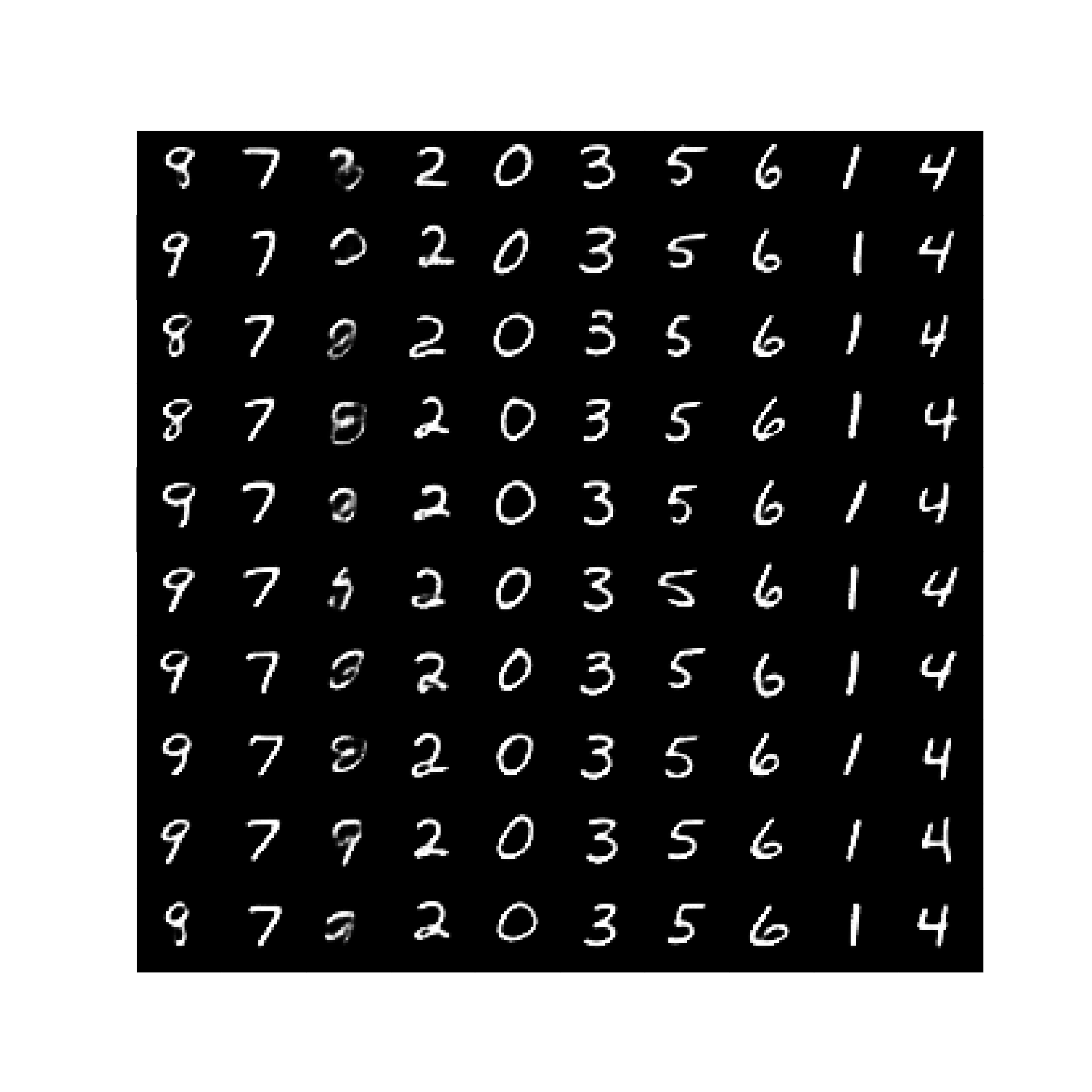}
    \caption{MultiMNIST}
    \label{fig:object_mnist}
  \end{subfigure}
  \begin{subfigure}{.45\textwidth}
    \centering
    \includegraphics[width=0.9\linewidth]{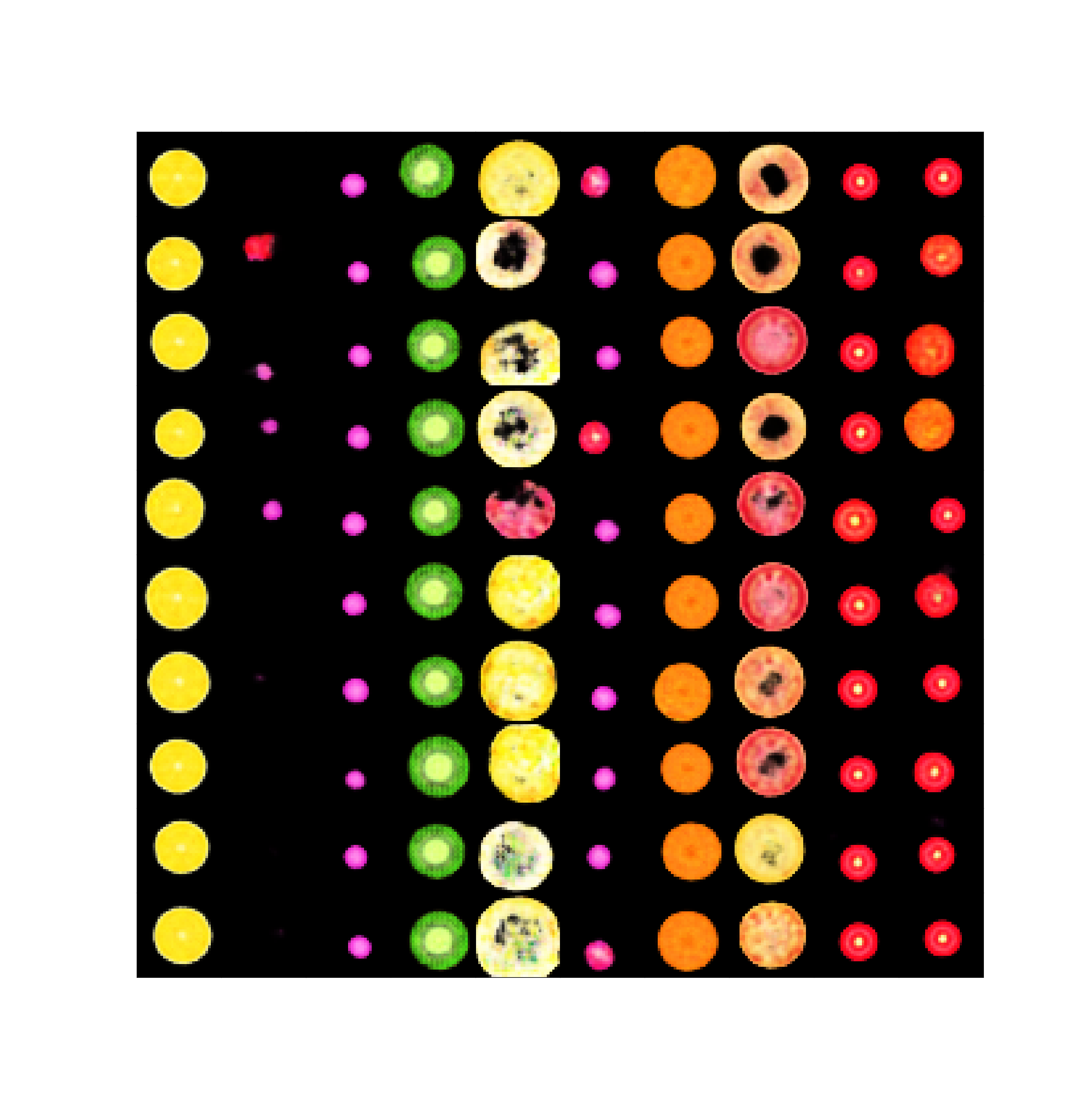}
    \caption{Fruit2D}
    \label{fig:object_fruit2d}
  \end{subfigure}
  \caption{Generated objects by varying $\zavg$ and $\zlocal$. The horizontal axis represents varying $\zavg$, and the vertical axis represents varying $\zlocal$, on both (a) and (b).}
  \label{fig:obj}
\end{figure}

It is expected that $\mu_k (1\leq k\leq C)$ represents the average feature of the $k$-th type of objects, and $\zwhat_{i}$ latent variable can be decomposed into:
\begin{equation}
  \zwhat_{i} = \zavg_{i} + \zlocal_{i}.
\end{equation}
$\zavg_{i} = \mu_k (1\leq k\leq C)$ if the $i$-th object is in the $k$-th category and $\zlocal$ represents the local feature of the object. By altering $\zavg$ or $\zlocal$, we should obtain new objects that belong to other categories or the same category with different styles, respectively.
In the experiment, we altered $\zavg$ and $\zlocal$ and observed the generated images for each object, as shown in Fig. \ref{fig:obj}. In Fig. \ref{fig:object_mnist}, objects in each cluster correspond to a type of digit, which is exactly what we expected (except for digit 8 in column 3). In Fig. \ref{fig:object_fruit2d}, categories with a large number of objects are grouped into multiple clusters, while categories with a small number are grouped into one cluster. This is due to the significant difference in number between various types. However, objects in a cluster come from a category in general.

\begin{figure}
  \begin{subfigure}{\textwidth}
    \centering
    \includegraphics[width=\linewidth]{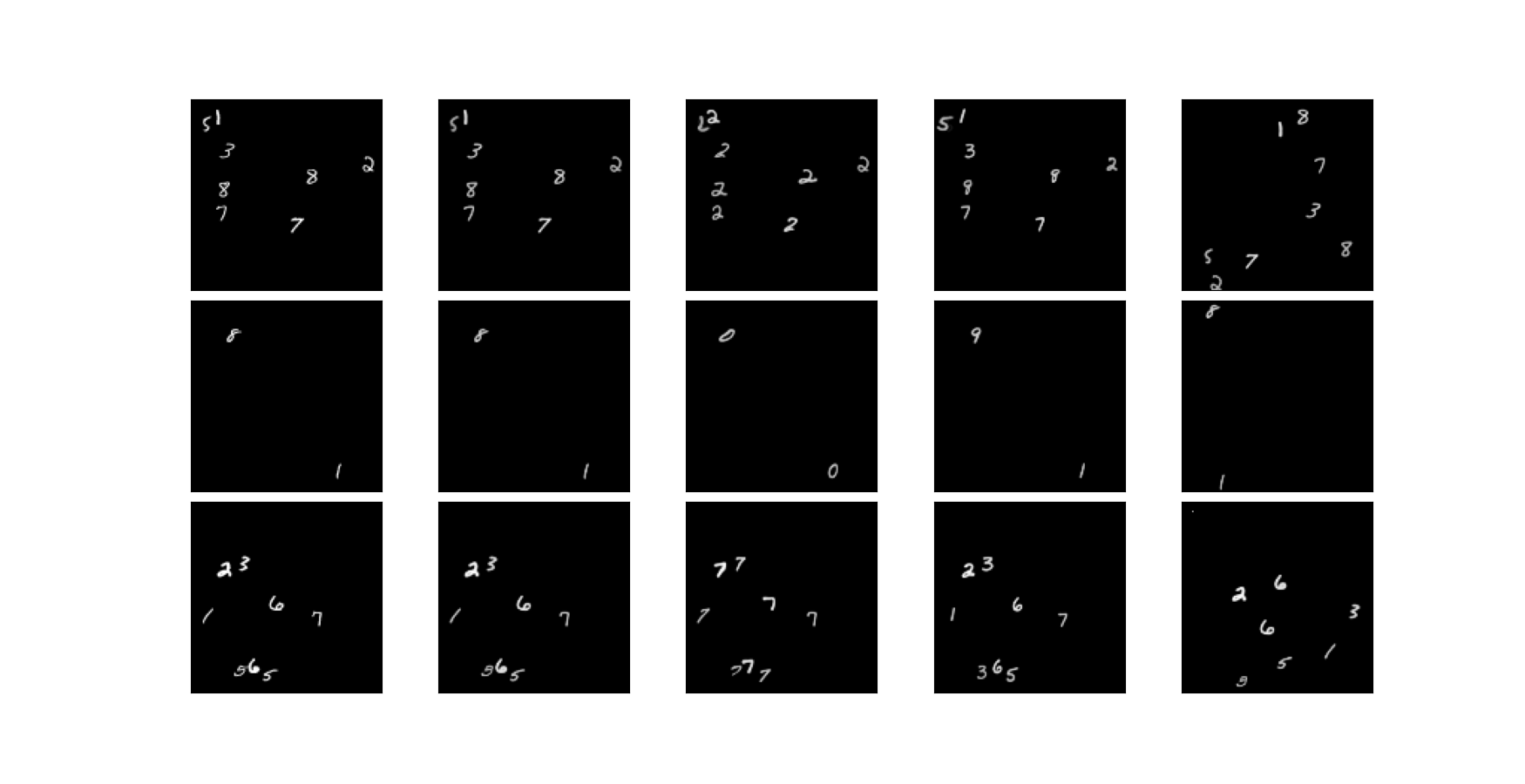}
    \caption{MultiMNIST}
    \label{fig:img_mnist}
  \end{subfigure}

  \begin{subfigure}{\textwidth}
    \centering
    \includegraphics[width=\linewidth]{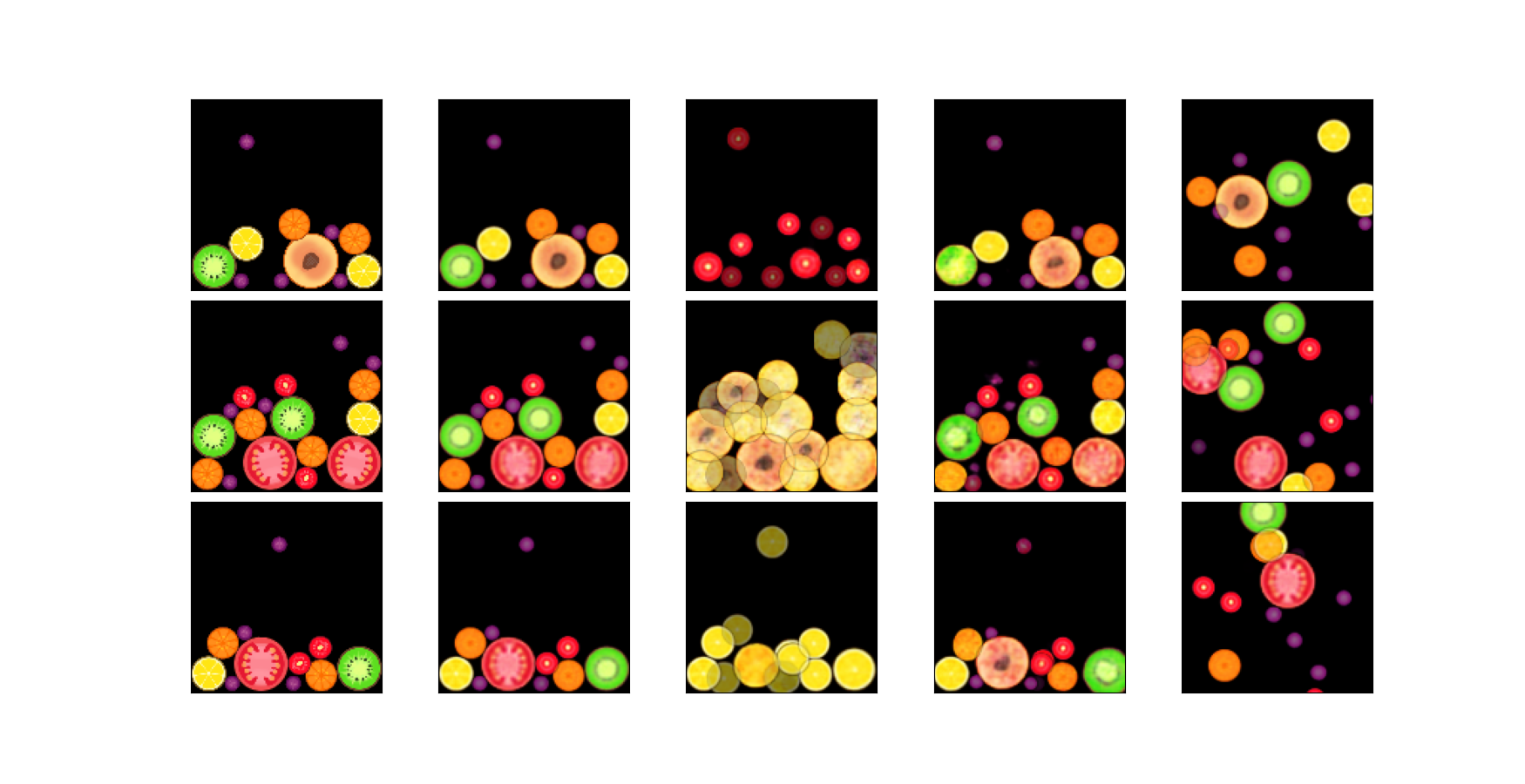}
    \caption{Fruit2D}
    \label{fig:img_fruit2d}
  \end{subfigure}

  \caption{Generated images by varying attributes and locations of objects. Columns 1 to 5 are numbered from left to right. Column 1 shows original images. Column 2 shows the generated images without varying $\zavg$, $\zlocal$, and $\zwhere$. Column 3 presents images generated by setting all $\zavg$ to the same random $\mu_k(1\leq k\leq C)$. Column 4 depicts images generated by varying $\zlocal$. Column 5 shows images generated by applying a random shuffle to $\boldz_i$.}
  \label{fig:gen_img}
\end{figure}

The structure of GMAIR ensures its ability to control object categories, object styles, and the positions of each object of the generated images by altering $\zavg$, $\zlocal$, and $\zwhere$. Examples are shown in Fig. \ref{fig:gen_img}.

This could provide a new approach for tasks such as style transfer, image generation, and data augmentation. Note that previous methods such as AIR, SPAIR, and its variants can also obtain similar results, but we achieve them in finer granularity.

\subsection{Quantitative Evaluations}
\label{sec:eval}
We quantitatively evaluate the models in terms of the AP of bounding boxes, ACC and NMI of the clusters, and the results are listed in Table \ref{table:res}. In the first part, we summarize some results of the state-of-the-art models for unsupervised clustering on MNIST dataset for comparison. In the second and the third part, we compare GMAIR to the state-of-the-art models for unsupervised object detection on MultiMNIST and Fruit2D dataset, respectively. The clustering results of SPAIR and SPACE are obtained by Gaussian mixture models (GMMs). Results show that GMAIR achieves competitive results on both clustering and localization.

\begin{table}
  \caption{Quantitative Results on Localization (AP) and Clustering (Accuracy and NMI)}
  \label{table:res}
  \centering
  \begin{tabular}{lllll}
    \toprule
    Model     & Dataset     & AP (\%, IoU=0.5) & ACC (\%) & NMI (\%)\\
    \midrule
    IIC & MNIST & --- & $98.4\pm 0.652$ & --- \\
    AAE (C=16) & MNIST & --- & $90.45\pm 2.05$ & --- \\
    AAE (C=30) & MNIST & --- & $95.90\pm 1.13$ & --- \\
    GMVAE (M=1) & MNIST & --- & $77.78\pm 5.75$ & --- \\
    GMVAE (M=10) & MNIST & --- & $82.31\pm 3.75$ & --- \\
    \midrule
    GMAIR & MultiMNIST & $97.3\pm 0.10$ & $80.4\pm 0.48$ & $75.5\pm 0.66$   \\
    SPAIR + GMM & MultiMNIST & $90.3$ & $59.4\pm 1.50$ & $56.3\pm 1.41$  \\
    SPACE + GMM & MultiMNIST & $96.7$ & $68.8\pm 3.43$ & $65.8\pm 2.85$ \\
    \midrule
    GMAIR & Fruit2D & $84.9\pm 1.56$ & $90.9\pm 0.32$ & $85.7\pm 1.25$   \\
    SPAIR + GMM & Fruit2D & $83.3$ & $88.1\pm 0.70$ & $78.4 \pm 0.51$    \\
    SPACE + GMM & Fruit2D & $93.8$ & $95.0\pm 1.99$ & $87.0\pm 2.20$ \\
    \bottomrule
  \end{tabular}
\end{table}

\section{Conclusion}
We introduce GMAIR, which combines spatial attention and a Gaussian mixture, such that it can locate and cluster unseen objects simultaneously. We analyze the ``what'' latent variables and clustering process, provide examples of GMAIR application to the task of image generation, and evaluate GMAIR quantitatively compared with SPAIR and SPACE.

\begin{ack}
This work was partially supported by the Research and Development Projects of Applied Technology of Inner Mongolia Autonomous Region, China under Grant No. 201802005, the Key Program of the National Natural Science Foundation of China under Grant No. 61932014, and Pudong New Area Science \& Technology Development Fundation under Grant No. PKX2019-R02. Yao Shen is the corresponding author.
\end{ack}

\newpage

\small
\bibliographystyle{abbrvnat}
\bibliography{references}

\normalsize
\newpage

\appendix

\section{Derivation of The KL Terms}
\label{appendix:elbo}

In this section, we derive the KL terms in Eqn. \ref{eqn:kl}. By assumption of $q(\boldz|\boldx)$ and $p(\boldz)$ (Eqn. \ref{eqn:qzx} and Eqn. \ref{eqn:pz}), we have:
\begin{equation}\label{eqn:kl2}
  \ep_{q(\boldz|\boldx)} \left[ \log(\frac{q(\boldz|\boldx)}{p(\boldz)}) \right] = \sum_{i=1}^{HW} \ep_{q(\boldz_i|\boldx)} \log \left( \frac{q(\boldz_i|\boldx)}{p(\boldz_i)} \right).
\end{equation}

The term $\ep_{q(\boldz_i|\boldx)} \log \left( \frac{q(\boldz_i|\boldx)}{p(\boldz_i)} \right)$ can further be expanded as follow:

\begin{align}\label{eqn:kl3}
  {} & \ep_{q(\boldz_i|\boldx)} \log \left( \frac{q(\boldz_i|\boldx)}{p(\boldz_i)} \right) \nonumber \\
  = & q(\zpres_i=0|\boldx)\log \left( \frac{q(\zpres_i=0|\boldx)}{p(\zpres_i=0)} \right) + q(\zpres_i=1|\boldx) \left( \log \left( \frac{q(\zpres_i=1|\boldx)}{p(\zpres_i=1)} \right) \right. \nonumber \\
   {} & \left.
   + \ep_{q(\zwhere_i,\zdepth_i,\zcat_i,\zwhat_i|\boldx)} \left[ \log \left( \frac{q(\zwhere_i,\zdepth_i, \zcat_i, \zwhat_i|\boldx)}{p(\zwhere_i,\zdepth_i, \zcat_i, \zwhat_i)} \right) \right] \right) \\
   = & KL \left(q(\zpres_i|\boldx) || p(\zpres_i) \right) \nonumber \\
   {} & + q(\zpres_i=1|\boldx) \ep_{q(\zwhere_i,\zdepth_i,\zcat_i,\zwhat_i|\boldx)} \left[ \log \left( \frac{q(\zwhere_i,\zdepth_i, \zcat_i, \zwhat_i|\boldx)}{p(\zwhere_i,\zdepth_i, \zcat_i, \zwhat_i)} \right) \right]. \nonumber
\end{align}

Continue to expand Eqn. \ref{eqn:kl3}:
\begin{align}\label{eqn:kl4}
  {} & \ep_{q(\zwhere_i,\zdepth_i,\zcat_i,\zwhat_i|\boldx)} \left[ \log \left( \frac{q(\zwhere_i,\zdepth_i, \zcat_i, \zwhat_i|\boldx)}{p(\zwhere_i,\zdepth_i, \zcat_i, \zwhat_i)} \right) \right] \nonumber \\
  = & \ep_{q(\zwhere_i|\boldx)} \log \left( \frac{q(\zwhere_i|\boldx)}{p(\zwhere_i)} \right) \nonumber \\
  + & \ep_{q(\zdepth_i|\boldx)} \log \left( \frac{q(\zdepth_i|\boldx)}{p(\zdepth_i)} \right) \\
  + & \ep_{q(\zcat_i,\zwhere_i|\boldx)} \log \left( \frac{q(\zcat_i|\boldx,\zwhere)}{p(\zcat_i)} \right) \nonumber \\
  + & \ep_{q(\zwhat_i,\zwhere_i,\zcat_i|\boldx)} \log \left( \frac{q(\zwhat_i|\boldx,\zwhere,\zcat_i)}{p(\zwhat_i|\zcat_i)} \right). \nonumber
\end{align}

By the definition of Kullback--Leibler divergence, the four terms in the RHS of Eqn. \ref{eqn:kl4} are indeed
\begin{align*}
  {} & KL \left( q(\zwhere_i|\boldx) || p(\zwhere_i) \right), \\
  {} & KL \left( q(\zdepth_i|\boldx) || p(\zdepth_i)
  \right), \\
  {} & \ep_{q(\zwhere_i|\boldx)} \left[ KL \left( q(\zcat_i|\boldx,\zwhere_i) || p(\zcat_i) \right) \right], \\
  \mbox{ and } & \ep_{q(\zwhere_i, \zcat_i|\boldx)} \left[ KL \left( q(\zwhat_i|\boldx,\zwhere_i,\zcat_i) || p(\zwhat_i|\zcat_i) \right) \right],
\end{align*}
respectively. Therefore, we complete the proof of Eqn. \ref{eqn:kl}.

During the implementation, we model discrete variables $\zpres$ and $\zcat$ using the Gumbel-Softmax approximation (\cite{jang2016categorical}). Therefore, all variables are differentiable using the reparameterization trick.

\section{Implementation Details}
\label{appendix:imp}

Our code is available at \url{https://github.com/EmoFuncs/GMAIR-pytorch}.

\subsection{Models}
Here, we describe the architecture of each module of GMAIR, as shown in Fig. \ref{fig:gmair}.
The backbone is a ResNet18 (\cite{he2016identity}) network with two deconvolution layers replacing the fully connected layer, as shown in Table \ref{table:backbone}.
Pres-head, depth-head, and where-head are convolutional networks that are only different from the number of output channels, as shown in Table \ref{table:heads}.
What-encoder and cat-encoder are multiple layer networks, as shown in Table \ref{table:encoders}.
Finally, the glimpse decoder is a deconvolutional network, as shown in Table \ref{table:decoder}.

For other models, we make use of code from \url{https://github.com/yonkshi/SPAIR_pytorch} for SPAIR, and \url{https://github.com/zhixuan-lin/SPACE} for SPACE. We utilize most of the default configuration for both models, and only change $A$ (the dimension of $\zwhat_i$) to $256$ for comparison, the size of the base bounding box to $72\times 72$ for large objects.

\subsection{Training and Hyperparameters}
The base set of hyperparameters for GMAIR is given in Table \ref{table:hypers}. The value $p$ (the prior on $\zpres$) drops gradually from $1$ to the final value $6\mbox{e-}6$, and the value $\alpha_{\mathsf{overlap}}$ drops from $2$ to $0$ in the early stage of training for stability. The learning rate is in the range of $[5\mbox{e-}5, 1\mbox{e-}4]$.

\subsection{Testing}
During testing phase, in order to obtain deterministic results, we use the value with the largest probability (density) for latent variables $\boldz$, instead of sampling them from the distributions. To be specific, we use $\boldsymbol{\pi}, \mu^{\mathsf{depth}}, p, \mu^{\mathsf{what}}, \mbox{ and } \mu^{\mathsf{where}}$ for $\zcat, \zdepth, \zpres, \zwhat$ and $\zwhere$, respectively.

\section{Calculation of AP, ACC and NMI}
\label{appendix:ap,acc,nmi}
The value of AP is calculated at threshold $\mbox{IoU}=0.5$ by using the calculation method from the VOC (\cite{everingham2010pascal}). Before calculating the ACC and NMI of clusters, we filter the incorrect bounding boxes. A predicted box $PB$ is correct iff there is a ground-truth box $GB$ such that $\operatorname{IoU}(PB, GB) > 0.5$, and the class of a correct predicted box $PB$ is assigned to the class of the ground-truth box $GB$ such that $\operatorname{IoU}(PB, GB)$ is maximized. After filtering, all correct predicted boxes are used for the calculation of ACC and NMI. Note that we still have many ways to assign each predicted category to a real category when calculating the value of ACC. In all of the ways, we select the one such that ACC is maximized, following \cite{dilokthanakul2016deep}. Formulas are shown in Eqn. \ref{eqn:acc} and Eqn. \ref{eqn:nmi} for the calculation of ACC and NMI:

\begin{equation}
  \label{eqn:acc}
  \mbox{ACC} = \frac{ \sum_{k=1}^{C} \max_{1\leq j\leq C'}| \left\{i| G_i=j, P_i=k \right\} |}{|P|}
\end{equation}

\begin{equation}
  \label{eqn:nmi}
  \mbox{NMI} = \frac{2 I(G, P)}{H(G) + H(P)}
\end{equation}

where $G$ and $P$ are respectively the ground-truth categories and predicted categories for all correct boxes, $C$ and $C'$ are the number of clusters and real classes, $H(\cdot)$ and $I(\cdot, \cdot)$ are the entropy and mutual information function, respectively.

\begin{table}
  \caption{Architecture of Backbone}
  \label{table:backbone}
  \centering
  \begin{tabular}{lllll}
    \toprule
    Layer & Type & Size & Act./Norm. & Output Size  \\
    \midrule
    resnet & ResNet18 (w/o fc) & {} & {} & $512\times 4\times 4$ \\
    deconv layer 1 & Deconv & $128$ & ReLU/BN & $128\times 8\times 8$ \\
    deconv layer 2 & Deconv & $64$ & ReLU/BN & $64\times 16\times 16$ \\
    \bottomrule
  \end{tabular}
\end{table}

\begin{table}
  \caption{Architectures of Pres/Depth/Where-Head}
  \label{table:heads}
  \centering
  \begin{tabular}{lllll}
    \toprule
    Layer & Type & Size & Act./Norm. & Output Size \\
    \midrule
    Input & {} & {} & {} & $64\times 16\times 16$ \\
    Hidden & Conv & $\left[ 3\times 3, 128 \right] \times 3$ & ReLU & $128\times 16\times 16$ \\
    Output & Conv & $1\times 1, 1/1/4$ & {} & $1/1/4 \times 16\times 16$ \\
    \bottomrule
  \end{tabular}
\end{table}

\begin{table}
  \caption{Architectures of What/Cat-Encoder}
  \label{table:encoders}
  \centering
  \begin{tabular}{lllll}
    \toprule
    Layer & Type & Size & Act./Norm. & Output Size   \\
    \midrule
    Input & Flatten & {} & {} & $(3\times 32\times 32=)3072$ \\
    Layer 1 & Linear & $3072 \times 128$ & ReLU & $128$ \\
    Layer 2 & Linear & $128 \times 256 $ & ReLU & $256$ \\
    Layer 3 & Linear & $256 \times 512 $ & ReLU & $512$ \\
    Output & Linear & $512 \times A/C$ & {} & $A/C$ \\
    \bottomrule
  \end{tabular}
\end{table}

\begin{table}
  \caption{Architecture of Glimpse-Decoder}
  \label{table:decoder}
  \centering
  \begin{tabular}{lllll}
    \toprule
    Layer & Type & Size & Act./Norm. & Output Size   \\
    \midrule
    Input & Linear & $A \times 256$ & ReLU & $256\times 1\times 1$ \\
    Layer 1 & Deconv & $128$ & ReLU/GN(8) & $128\times 2\times 2$ \\
    Layer 2 & Deconv & $128$ & ReLU/GN(8) & $128\times 4\times 4$ \\
    Layer 3 & Deconv & $64$ & ReLU/GN(8) & $64\times 8\times 8$ \\
    Layer 4 & Deconv & $32$ & ReLU/GN(8) & $32\times 16\times 16$ \\
    Conv & Conv & $3\times 3, 32$ & ReLU/GN(8) & $32\times 16\times 16$ \\
    Layer 5 & DeConv & $16$ & ReLU/GN(4) & $16\times 32\times 32$ \\
    Output & Conv & $3 \times 3, 3$ & {} & $3\times 32\times 32$ \\
    \bottomrule
  \end{tabular}
\end{table}
\begin{table}
  \caption{Base Hyperparameters}
  \label{table:hypers}
  \centering
  \begin{tabular}{lllll}
    \toprule
    Description & Variable & Value \\
    \midrule
    Base bbox size & $(a_h, a_w)$ & $(72, 72)$ \\
    Batch size & {} & $16$ \\
    Dim. of $\zwhat_i$ & $A$ & $256$ \\
    Dim. of $\zcat_i$ & $C$ & $10$ \\
    Glimpse size & $(H_{obj}, W_{obj})$ & $(32, 32)$ \\
    Learning rate & {} & $[5\mbox{e-}5, 1\mbox{e-}4]$ \\
    Loss Coef. of $L_{\mathsf{cat}}$ & $\alpha_{\mathsf{cat}}$ & 1 \\

    Loss Coef. of $L_{\mathsf{overlap}}$ & $\alpha_{\mathsf{overlap}}$ & $2\rightarrow 0$ \\
Loss Coef. of $L_{\mathsf{depth}}$ & $\alpha_{\mathsf{depth}}$ & 1 \\
    Loss Coef. of $L_{\mathsf{pres}}$ & $\alpha_{\mathsf{pres}}$ & 1 \\
    Loss Coef. of $L_{\mathsf{recon}}$ & $\alpha_{\mathsf{recon}}$ & 8,16 \\
    Loss Coef. of $L_{\mathsf{what}}$ & $\alpha_{\mathsf{recon}}$ & 1 \\
    Loss Coef. of $L_{\mathsf{where}}$ & $\alpha_{\mathsf{recon}}$ & 1 \\

    Prior on $\zcat$ & $\boldsymbol{\pi}$ & $(1/C,...,1/C)$ \\
    Prior on $\zdepth$ & $(\mu^{\mathsf{depth}}_{\mathsf{prior}}, {\sigma^{\mathsf{depth}}_{\mathsf{prior}}})$ & $(0, 1)$ \\
    Prior on $\zpres$ & $p$ & $1\rightarrow 6\mbox{e-}6$ \\
    Prior on $\zwhere$ & $(\mu^{\mathsf{where}}_{\mathsf{prior}}, {\sigma^{\mathsf{where}}_{\mathsf{prior}}})$ & $(0, 1)$ \\
    Prior on $\zwhat$ & $(\mu^{\mathsf{what}}_{\mathsf{prior}}, {\sigma^{\mathsf{what}}_{\mathsf{prior}}})$ & $(0, 1)$ \\

    \bottomrule
  \end{tabular}
\end{table}

\section{Additional Experiment Results}
\label{appendix:exps}

\begin{figure}
  \begin{subfigure}{.24\textwidth}
    \centering
    \includegraphics[width=\linewidth]{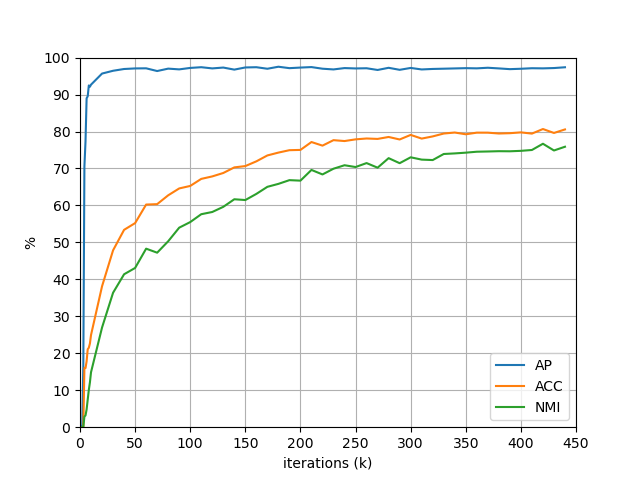}
    \caption{AP(IoU=0.5), ACC and NMI during training}
    \label{fig:what-ap-full}
  \end{subfigure}
  \begin{subfigure}{.24\textwidth}
    \centering
    \includegraphics[width=\linewidth]{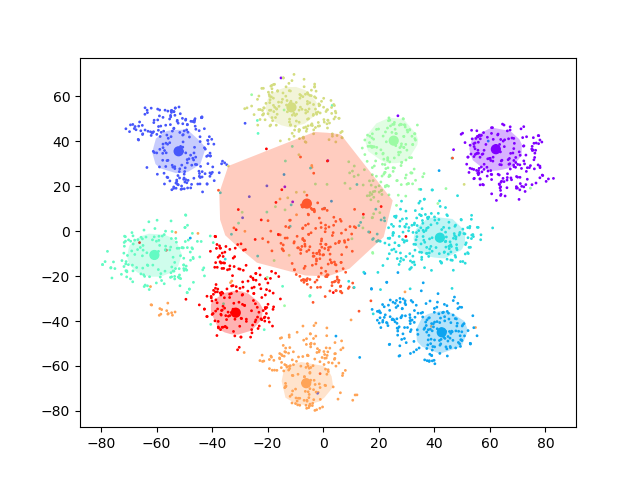}
    \caption{``What'' latent space, at 220k iterations}
    \label{fig:what-220}
  \end{subfigure}
  \begin{subfigure}{.24\textwidth}
    \centering
    \includegraphics[width=\linewidth]{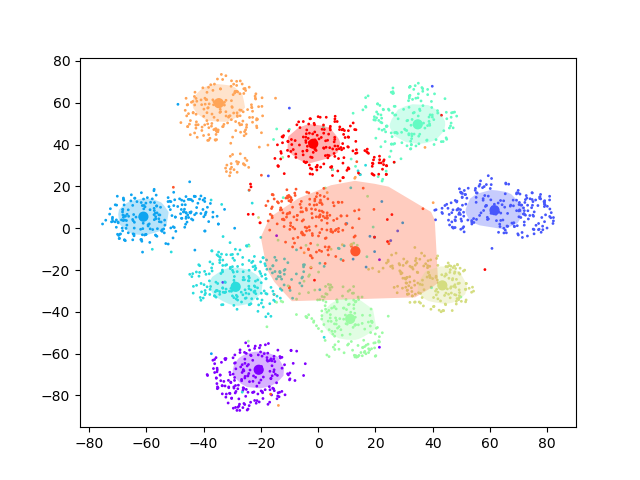}
    \caption{``What'' latent space, at 330k iterations}
    \label{fig:what-330}
  \end{subfigure}
  \begin{subfigure}{.24\textwidth}
    \centering
    \includegraphics[width=\linewidth]{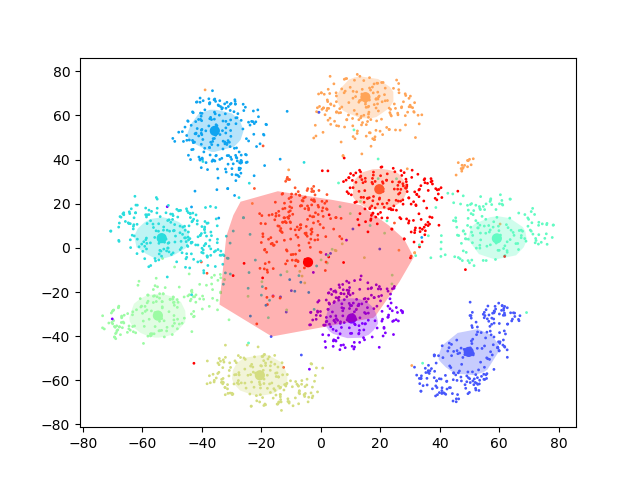}
    \caption{``What'' latent space, at 440k iterations}
    \label{fig:what-440}
  \end{subfigure}
  \newline
  \begin{subfigure}{.24\textwidth}
    \centering
    \includegraphics[width=.75\linewidth]{data_x.png}
    \caption{Original image}
    \label{fig:what-image-ori-2}
  \end{subfigure}
  \begin{subfigure}{.24\textwidth}
    \centering
    \includegraphics[width=.75\linewidth]{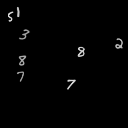}
    \caption{Generated image, at 220k iterations}
    \label{fig:what-image-220}
  \end{subfigure}
  \begin{subfigure}{.24\textwidth}
    \centering
    \includegraphics[width=.75\linewidth]{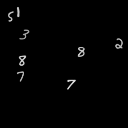}
    \caption{Generated image, at 330k iterations}
    \label{fig:what-image-330}
  \end{subfigure}
  \begin{subfigure}{.24\textwidth}
    \centering
    \includegraphics[width=.75\linewidth]{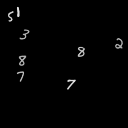}
    \caption{Generated image, at 440k iterations}
    \label{fig:what-image-440}
  \end{subfigure}
  \caption{``What'' representation and cluster analysis after 100k iterations.
  (a) Average precision (AP), accuracy (ACC), normalized mutual information (NMI) during training.
  (b-d) Visualized ``what'' latent space by t-SNE (\cite{van2008visualizing}) at 220k, 330k, and 440k iterations, respectively.
  (e) Sample of original image.
  (f-h) Samples of generated image at 220k, 330k, and 440k iterations, respectively.}
  \label{fig:what-full}
\end{figure}

The graphs of ``what'' representation after 100k iterations are shown in Fig. \ref{fig:what-full}.

\end{document}